\documentclass[]{article}
\usepackage[margin=1in]{geometry}
\usepackage[comma]{natbib}
\usepackage{url}
\usepackage{graphicx}
\usepackage{comment}
\title{Interpretable Machine Learning for Discovery: Statistical Challenges \& Opportunities}

\author{Genevera I. Allen$^{1,2}$, Luqin Gan$^{2}$, and Lili Zheng$^{3}$}
\begin{document}

\maketitle

\begin{abstract}
New technologies have led to vast troves of large and complex datasets across many scientific domains and industries.  People routinely use machine learning techniques to not only process, visualize, and make predictions from this big data, but also to make data-driven discoveries.  These discoveries are often made using Interpretable Machine Learning, or machine learning models and techniques that yield human understandable insights.  In this paper, we discuss and review the field of interpretable machine learning, focusing especially on the techniques as they are often employed to generate new knowledge or make discoveries from large data sets.  We outline the types of discoveries that can be made using Interpretable Machine Learning in both supervised and unsupervised settings.  Additionally, we focus on the grand challenge of how to validate these discoveries in a data-driven manner, which promotes trust in machine learning systems and reproducibility in science.  We discuss validation from both a practical perspective, reviewing approaches based on data-splitting and stability, as well as from a theoretical perspective, reviewing statistical results on model selection consistency and uncertainty quantification via statistical inference.  Finally, we conclude by highlighting open challenges in using interpretable machine learning techniques to make discoveries, including gaps between theory and practice for validating data-driven-discoveries.  
\end{abstract}


\footnotetext[1]{Departments of Electrical and Computer Engineering and Computer Science, Rice University, Houston, TX, USA 77005; Neurological Research Institute, Baylor College of Medicine, Houston, TX, USA 77030; Email: gallen@rice.edu}
\footnotetext[2]{Department of Statistics, Rice University, Houston, TX, USA 77005.}
\footnotetext[3]{Department of Electrical and Computer Engineering, Rice University, Houston, TX, USA 77005.}

\section{Introduction}

Machine learning systems have gained widespread use in science, technology, and society.  Given the increasing number of high-stakes machine learning applications and the growing complexity of machine learning models, many have advocated for interpretability and explainability to promote understanding and trust in machine learning results \citep{rasheed2022explainable,toreini2020relationship,broderick2023toward}. In response, there has been a recent explosion of research on Interpretable Machine Learning (IML), mostly focusing on new techniques to interpret black-box systems; see \cite{molnar2020interpretable,lipton2018mythos,guidotti2018survey,doshi2017towards,du2019techniques,murdoch2019definitions,carvalho2019machine} for recent reviews of the IML and explainable artificial intelligence literature.  While most of these interpretability techniques were not necessarily designed for this purpose, they are increasingly being used to mine large and complex data sets to generate new insights \citep{roscher2020explainable}. These so-called data-driven discoveries are especially important to advance data-rich fields in science, technology, and medicine. While prior reviews focus mainly on IML techniques, we primarily review how IML methods promote data-driven discoveries, challenges associated with this task, and related new research opportunities at the intersection of machine learning and statistics. 

In the sciences and beyond, IML techniques are routinely employed to make new discoveries from large and complex data sets; to motivate our review on this topic, we highlight several examples.  First, feature importance and feature selection in supervised learning are popular forms of interpretation that have led to major discoveries like discovering new genomic biomarkers of diseases \citep{guyon2002gene}, discovering physical laws governing dynamical systems \citep{brunton2016discovering}, and discovering lesions and other abnormalities in radiology \citep{borjali2020deep,reyes2020interpretability}.  While most of the IML literature focuses on supervised learning \citep{molnar2020interpretable,lipton2018mythos,guidotti2018survey,doshi2017towards}, there have been many major scientific discoveries made via unsupervised techniques and we argue that these approaches should be included in any discussion of IML.  For example, one of the earliest and most important machine learning findings in medicine was the discovery of genomic subtypes of breast cancer using hierarchical clustering of gene expression data \citep{perou2000molecular}, which led to new ways to diagnose and treat cancer based on a patient's specific genomic subtype and ushered in an era of personalized medicine \citep{hassan2022innovations}.  Clustering techniques have also been used to discover galaxies from astronomical surveys \citep{materne1978structure} and discover communities with similar political affiliations \citep{ozer2016community}.  Other major unsupervised discoveries include discovering major climate patterns like EL Nino and their localized effects via dimension reduction \citep{houghton2020nino} and discovering the functional organization of the brain via network models \citep{rubinov2010complex}. These are just a few of many examples of how IML techniques have led to new scientific discoveries. As the size and complexity of scientific data continues to grow, IML techniques will be ever more valuable for mining this data to generate new findings and advance science, hence motivating our review on this topic.

In this article, we review IML for the purpose of generating new data-driven discoveries. We also discuss several challenges that come with using IML for discovery, review statistical and other research that has sought to address these challenges, and highlight many associated open research opportunities. We organize this article by first reviewing the extensive IML literature in Section~\ref{sec:IML}. Next in Section~\ref{sec:IML_types}, we review IML techniques, but instead of organizing this according to technique-type as in most other IML reviews, we discuss IML techniques as they are used to generate different discovery types.  Our discussion includes both supervised and unsupervised techniques, given the importance of the latter for making discoveries.  In order for IML findings to lead to accurate discoveries, however, we need them to be replicable and reliable \citep{yu2020veridical}, which also promotes trust in machine learning results \citep{rasheed2022explainable,toreini2020relationship,broderick2023toward}.  In other words, we need approaches to validate IML discoveries.  But unfortunately, validation for IML is not widely discussed or applied in practice as it presents many more challenges than validating ML predictions. In Section~\ref{sec:validation}, we discuss the grand challenge of validating IML discoveries and review several practical validation strategies with examples.  Then in Section~\ref{sec:theory_inference}, we approach validation from a theoretical perspective and review statistical theory and statistical inference approaches that can help determine when IML techniques will find the desired discovery with high probability (Section~\ref{sec:theory}) as well as help quantify the uncertainty in IML discoveries via confidence intervals and statistical hypothesis testing (Section~\ref{sec:inference}).  We finally conclude with a discussion of the major open problems and opportunities in IML for discovery in Section~\ref{sec:disc}.

\section{Interpretable Machine Learning: Definitions, Rationale, \& Categories}\label{sec:IML}

\begin{figure}
    \centering
    \includegraphics[width=3.5in]{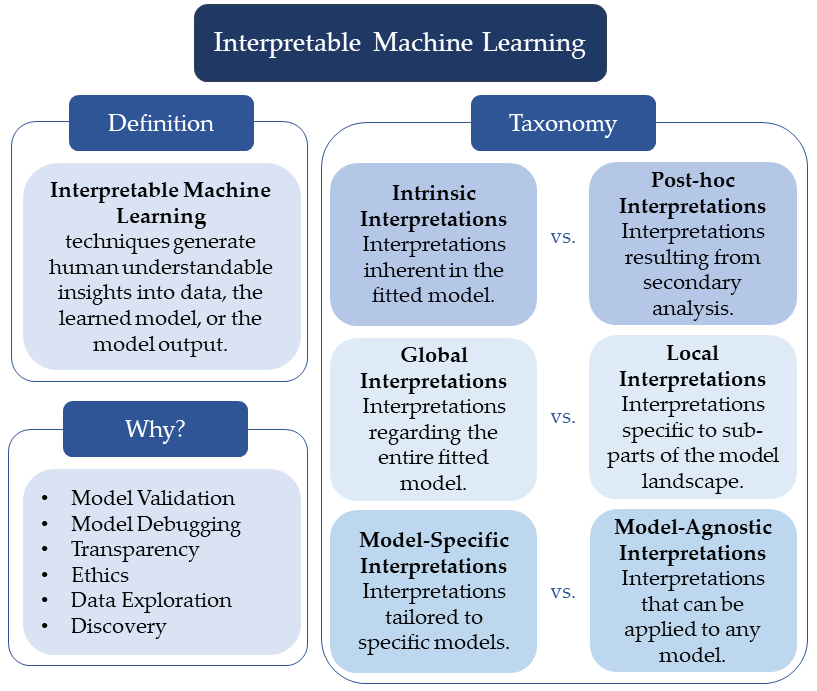}
    \caption{Overview of Interpretable Machine Learning.}
    \label{fig:overview}
\end{figure}

Before focusing on IML for making discoveries, we review the growing literature on IML.  We discuss definitions, reasons for using IML, and taxonomies that provide a systematic way to describe IML techniques.  These are summarized in Figure~\ref{fig:overview}.

\subsection{What is Interpretable Machine Learning?}
Many have discussed IML, yet there is not a universally accepted consensus definition \citep{murdoch2019definitions,roscher2020explainable,du2019techniques,rudin2014algorithms,arrieta2020explainable}. Imprecise definitions have likely led to a lack of consensus on how to study and validate IML techniques, a major concern when these methods are used to make data-driven discoveries \citep{rudin2022interpretable,gilpin2018explaining}.  We adopt a broad definition of interpretable machine learning: {\it {\bf Interpretable machine learning} is the use of machine learning techniques to generate human-understandable insights into data, the learned model, or the model output.} In other words, interpretable machine learning is very general and provides an understanding of any aspect of the machine learning process: the model inputs (data), the model insides or model guts (the model parameters or learned model, or even how the model interacts with data), and the model outputs (predictions or decisions based on the data and model).  As many have noted, what is considered a human-understandable insight depends on the intended audience and the domain area; thus, interpretations in machine learning are domain, problem, and audience specific \citep{murdoch2019definitions,roscher2020explainable}. 

\subsection{Why Interpretability?} 

Why do we need interpretability in machine learning?  Many have proposed a number of reasons and uses for interpretable machine learning \citep{du2019techniques,murdoch2019definitions,molnar2020interpretable,lipton2018mythos,guidotti2018survey,carvalho2019machine,doshi2017towards,roscher2020explainable} which we briefly review here.    

{\it Model Validation}.  When fitting complex machine learning systems, the modeler may need to check that the model is performing and behaving in the desired manner, or perform model validation.    One may ask: Does this model make sense?  Is this model consistent with my prior expectations or knowledge about the system?  This form of human validation requires human interpretable machine learning models. 

{\it Model Debugging}.  When something goes wrong in a machine learning system, how can one diagnose a problem in a machine learning system if they don't understand the model and how it interacts with the data?  Interpreting and understanding machine learning systems are critical for diagnosing, debugging and fixing systems \citep{koh2017understanding}. 

{\it Transparency, Accountability \& Trust}. Interpretable machine learning approaches often help to make black-box and other machine learning systems easier for humans to understand and hence more transparent.  This transparency is critical for promoting accountability and trust of machine learning systems which are necessary for their utilization in high-stakes societal applications \citep{rudin2019stop,samek2019towards,xu2019explainable}.   

{\it Ethics}.  There has been an increasing focus on ensuring that machine learning algorithms are fair and ethical \citep{doshi2017towards}. Due to biases that exist in our society, machine learning algorithms that are trained on possibly biased data can often exacerbate these biases leading to unfair predictions that are discriminatory \citep{guidotti2018survey}. Understandable machine learning techniques are needed to both assess and improve the fairness of machine learning in critical societal applications. 

{\it Data Exploration}.  John Tukey coined the term Exploratory Data Analysis and promoted this as the critical first stage of data analysis \citep{tukey1977exploratory}. Human-interpretable techniques can help gain insights into major patterns, trends, groups, or artifacts of the data. These data exploration insights are then used to clean and prepare data for modeling, make downstream modeling decisions, and visualize and interpret model outputs \citep{murdoch2019definitions,berkhin2006survey}.  

{\it Discovery}. As data has grown in size and complexity, we often rely on machine learning techniques to make discoveries, or in other words, find rare signals in a sea of data.  Using interpretable machine learning techniques to make data-driven discoveries is the main focus of this review.

\subsection{A Taxonomy of IML Techniques}\label{sec:tax}

Recently, many have discussed interpretable machine learning techniques and proposed various categorizations to systematize discussion and evaluation of the approaches  \citep{molnar2020interpretable,lipton2018mythos,guidotti2018survey,doshi2017towards}.  While there is not complete agreement in the literature on these categories, we discuss three main dimensions or axes along which most interpretable machine learning techniques lie and give examples of methods falling under each designation.  We also discuss how these categories of techniques relate to the task of using IML methods for generating new discoveries.

\subsubsection{Intrinsic vs. Post-hoc Interpretability}

A major axis that differentiates IML techniques is intrinsic versus post-hoc interpretability. Intrinsic interpretations are understandings that are inherent in the fitted model itself. In other words, the user needs to simply fit a model to produce the desired interpretation.  Examples include trees, additive models, or regularization approaches which make the fitted model more understandable by adding constraints like sparsity or smoothness. More recently in deep learning, many have proposed models that are more intrinsically interpretable by constraining the final layer in a deep neural network to follow certain prototypes or interpretability constraints \citep{dong2017towards,rudin2019stop}. In contrast, post-hoc interpretations require a secondary technique to be applied to the fitted model or model outputs for the sole purpose of interpretation. Examples of post-hoc interpretations include backpropagation-related methods which traverse the learned neural network architecture to assign importance scores to each feature and local interpretable model explanations (LIME) which fits a second, simple, and interpretable model approximating the black-box model at a particular input \citep{molnar2020interpretable}. Additionally, most supervised model-agnostic interpretations, discussed subsequently, are post-hoc in nature. Very little attention has been paid to unsupervised learning techniques in the context of IML.  But, we argue that all unsupervised learning techniques are naturally intrinsically interpretable as their objective is to find some meaningful structure that helps the user gain insights into the data, hence falling under our definition of IML. One can still use post-hoc interpretations of unsupervised findings, however. Consider that after clustering, one may perform a secondary analysis to determine which features are most responsible for separating the clusters \citep{satija2015spatial}.

Many have argued that intrinsic interpretations are preferable to post-hoc interpretations \citep{lipton2018mythos,varshney2019trustworthy,rudin2019stop}.  For the purpose of making data-driven discoveries, however, there is not a particular preference as long as the interpretations accurately capture the discovery of interest.  For intrinsic interpretations, this means the model must fit the data well and closely approximate the true generating model for the interpretations to reflect true discoveries.  In linear regression, for example, the intrinsic interpretation of feature importance based on estimated coefficients will only be accurate if the true underlying model is linear or approximately linear.  For post-hoc interpretations, on the other hand, both the original model and the secondary analysis must accurately capture the data-generating process to yield accurate interpretations.  If a deep learning model fits the data well, but a secondary analysis with LIME does not sufficiently capture the original deep learning model, then interpretations and resulting discoveries will not be accurate \citep{zhang2019should}.

\subsubsection{Model-Specific vs. Model-Agnostic Interpretations}

Another dimension along which we can categorize IML techniques is by whether they are model-specific or model-agnostic.  Model-specific interpretations are tailored to the model and cannot generalize across models. Model-agnostic interpretations can be applied to any model and interpreted in a similar manner for all models.  
There are many forms of model-specific feature importance; these include coefficients in generalized linear or additive models, feature importance scores for trees, or the plethora of deep learning specific techniques for feature attribution like backpropagation methods \citep{molnar2020interpretable}. 
On the other hand, there are several model-agnostic feature importance methods that can be used for any supervised model; these include Shapley values, feature permutations, feature occlusion, and LIME \citep{molnar2020interpretable}.
Note that model-specific interpretations are not necessarily intrinsic interpretations; consider that feature importance scores for trees and guided backpropagation feature attribution are both model-specific but post-hoc. In contrast, most model-agnostic interpretations are post-hoc in nature.

For the same reason that many prefer intrinsic interpretations to post-hoc interpretations, many have argued that model-specific interpretations are preferable \citep{ribeiro2016should,lundberg2017unified}. 
Yet, we point out that there are several advantages to model-agnostic interpretations. Importantly, model-agnostic interpretations can be understood in the same way across all models.  This is particularly useful for model comparisons; for feature importance, for example, it allows one to compare importance scores for each feature derived from several model families.  We argue that this advantage is especially important for the task of making data-driven discoveries where validating the interpretations is critical. One easy approach is to try many different model families and check if the interpretations are the same across all of these models.  This type of validation is easier with model-agnostic interpretations that can be directly compared.  On the other hand, it is typically easier to study model-specific interpretations theoretically to understand under what conditions the resulting discoveries accurately recover some aspect of the true model, a topic we will discuss further in Section 5.

\subsubsection{Global vs. Local Interpretations}

A final major dimension along which we can categorize IML techniques is based on whether the approach offers a local or a global interpretation.  Global interpretations reveal the overall structure of the fitted model.  On the other hand, local interpretations only yield model insights based on sub-parts of the model input space; these could include local interpretations about a single observation or a subset of the domain.  To make these distinctions concrete, again consider the example of feature importance in supervised learning.  Here, methods previously mentioned like coefficients in linear or additive models, tree-based feature importance, and backpropagation-based feature attribution are all global interpretations that capture the relevance of each feature for all model predictions.  In contrast, methods like LIME and saliency maps, 
highlight the important features of a single new test instance or observation \citep{ribeiro2016should,molnar2020interpretable}. Similarly, in unsupervised learning, consider the task of dimension reduction. Methods like principal components analysis and spectral embedding yield global interpretations, revealing global patterns represented in all observations in each of the factors. In contrast, local embedding and neighborhood embedding methods, like t-SNE and UMAP, highlight local interpretations through patterns and relationships amongst particular neighborhoods.

When using interpretable machine learning to make discoveries, global interpretations are more commonly employed as they reveal discoveries reflective of all the input data and model landscape.  Yet, local interpretations are increasingly important to make discoveries amongst subgroups of observations.  An example application of this is in precision medicine where we might seek to discover important genomic biomarkers for each individual patient or sub-groups of similar patients.

\section{Types of IML Discoveries \& Techniques}\label{sec:IML_types}

Recent research in IML has produced an abundance of interpretability techniques, as thoroughly reviewed in \cite{molnar2020interpretable}. But these works focus on the types of techniques and not the types of data-driven discoveries that various techniques can make.  Here, we organize this section to highlight the major types of discoveries achieved through interpretations of machine learning models. Importantly and distinct from the IML literature, we place great emphasis on unsupervised techniques, which are popularly used throughout the sciences to make discoveries from unlabeled data.

\begin{figure}
    \centering
    \includegraphics[width=\textwidth]{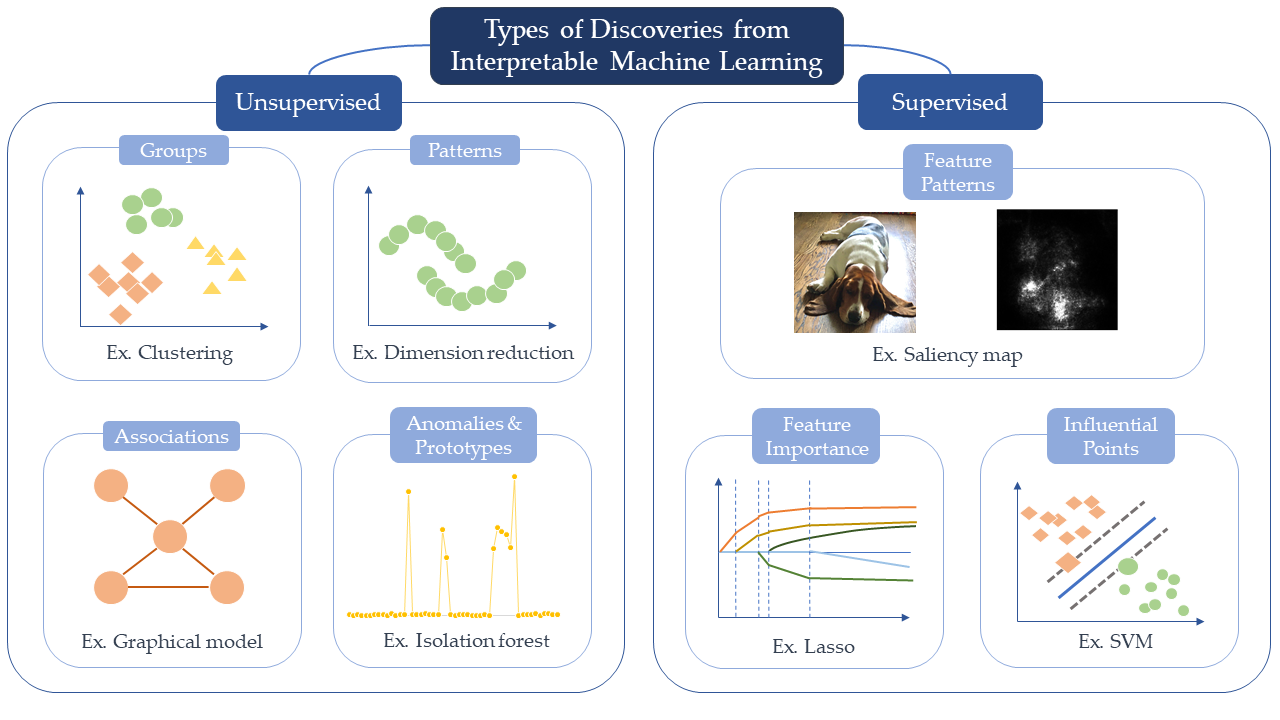}
    \caption{Overview of the broad types of unsupervised and supervised data-driven discoveries that can be made using interpretable machine learning techniques.}
    \label{fig:discovery}
\end{figure}

\subsection{Unsupervised Discoveries}

Most work on IML has focused on supervised models and interpreting the results of predictive systems \citep{rudin2022interpretable}. In scientific domains, however, some of the most widespread uses of machine learning are in unsupervised settings;  hence in this section, we review major types of unsupervised discoveries and highlight which types of IML techniques are employed to generate these discoveries.

\subsubsection{Groups}
Uncovering hidden group structures in large datasets is a common and popular type of unsupervised discovery. There are many well-established clustering techniques used for this task including $K$-means, hierarchical clustering, mixture modeling, and spectral clustering, among many others \citep{Hennig2015HandbookOC}. Beyond group membership, other types of interpretations related to clustering include uncovering groups of both observations and features simultaneously via biclustering, discovering nested group structure via hierarchical clustering, detecting localized important regions via spatial clustering, and finding a subset of features that distinguish groups of observations via sparse clustering \citep{witten2010framework}. Clustering has been applied broadly and is a nearly ubiquitous technique in unsupervised and exploratory analysis; groups found via clustering have also led to several major scientific discoveries such as finding gene expression patterns and/or genomic subtypes of diseases like cancer \citep{perou2000molecular}. 

\subsubsection{Patterns \& Trends}

When conducting unsupervised analyses, a typical first task is to visualize and explore the data to look for major patterns and trends. Often, important unsupervised discoveries can be made through these visual inspections of the data.  For large multivariate data, dimension reduction approaches reduce the data down to a smaller number of components that retain important structure, or patterns, in the data. There are a plethora of dimension reduction techniques including linear approaches like principal components analysis (PCA), non-negative matrix factorization and independent component analysis, or non-linear approaches like spectral embedding, multi-dimensional scaling, isomap, t-SNE, UMAP, or autoencoders; see \cite{fodor2002survey} for a recent review of such approaches. Each of these approaches is optimized to find slightly different types of patterns. For example, PCA finds variance-maximizing patterns that preserve the global structure whereas t-SNE finds localized patterns that preserve neighborhood and group structure.  

\subsubsection{Associations}

Discovering associations, or important relationships amongst features, is another widely used type of unsupervised discovery. Most typically find linear or nonlinear associations by exploring all possible pairwise interactions amongst features, using correlation, mutual information, or other such metrics. Recently, there has been a surge of interest to explore feature relationships using graphical models \citep{lauritzen1996graphical}.  In Markov Networks, or undirected graphical models, for example, the goal of structural learning is to estimate conditional dependencies between features; structural learning in Bayesian Networks or Directed Acyclic Graphs (DAG) seeks to learn directed relationships and is an important part of causal discovery \citep{drton2017structure}.

\subsubsection{Anomalies \& Prototypes}

Other types of unsupervised discoveries that are perhaps less commonly used are finding anomalies (rare entities) or prototypes (typical entities). Anomalies are rare but noteworthy observations.  Techniques for anomaly detection are similar to those for outlier detection; these include distance-based approaches which often employ dimension reduction, clustering approaches like single-linkage hierarchical clustering or DBSCAN, the one-class support vector machine, or the isolation forest \citep{hodge2004survey}. 
Alternatively, sometimes one seeks to find the most representative observations, or prototypes. Adaptions of other unsupervised approaches, especially dimension reduction and clustering, are typically employed for this task \citep{bien2011prototype}. 

\subsection{Supervised Discoveries}

Supervised learning has been the focus of the vast majority of the interpretable machine learning literature \citep{molnar2020interpretable,lipton2018mythos,guidotti2018survey,doshi2017towards}. This occurs as some of the best performing predictive models, such as deep learning and tree-based ensembles, are essentially black-boxes that are not intrinsically interpretable and difficult to decipher. Thus, interpretations of these predictive models are critical for generating new insights and making data-driven discoveries. 

\subsubsection{Feature Importance \& Feature Selection}

 Perhaps the most common and popular form of interpretation in supervised models is understanding how each feature influences a model's predictions, often referred to as feature importance. Related to this is feature selection which finds the best subset of features that maximize predictive accuracy.  Importantly, feature importance and feature selection in supervised learning offers a form of multivariate or conditional feature interpretation: given all other features in the model, what is the added benefit of including a particular feature? This conditional feature interpretation is much stronger than marginally assessing how each feature relates to an outcome and has been used extensively to discover important features. 
Let us review the many types of methods for interpreting features in supervised learning through the context of our IML taxonomies from Section~\ref{sec:tax}. First, consider global and model-specific feature importance metrics. For linear or generalized additive models, the feature weight (or parameter or coefficient) can be directly interpreted as the conditional feature importance, offering intrinsic feature interpretability. Tree-based ensembles offer post-hoc interpretability by the feature importance scores based on loss difference for each split. 
In deep learning, post-hoc approaches are popular and include several feature importance scores calculated via gradient-based methods which traverse the fitted neural network to attribute relevance to each input feature \citep{samek2021explaining}.
Local, post-hoc, and model-specific methods are popular in computer vision where measures such as saliency maps and GradCAM highlight which pixels in a specific image were used to generate the predicted label \citep{samek2021explaining}. 
There are also several model-agnostic metrics that can be used with any supervised learning model, including methods that yield global interpretations, like feature occlusion, feature permutation, and Shapley values, as well as local interpretations such as LIME. See \cite{molnar2020interpretable} on these methods.  
There is an equally impressive literature on feature selection for supervised learning; most of these strategies offer intrinsic and model-specific interpretations by working with the empirical risk minimizer or loss function. As finding the best subset of features is a combinatorially hard optimization problem, people typically turn to greedy step-wise methods, like Recursive Feature Elimination, or regularization strategies that relax the best subset constraint.  Popular approaches to the latter include the $\ell_{1}$ or the Lasso penalty that encourages sparsity in the feature weights \citep{tibshirani1996regression}.  The Lasso and other regularization approaches are routinely employed across all areas of machine learning to aid in interpreting features \citep{li2020survey}.  

\subsubsection{Feature Interactions \& Feature Representations}
Beyond the importance of each individual feature, one may want to understand higher-order interactions or feature patterns that are important for a model's predictions.   Decision trees and their extensions offer natural ways of assessing model-specific feature interactions, but there are several model agnostic approaches such as Friedman's H-statistic, variable interaction networks, and partial dependence functions; see \cite{molnar2020interpretable} for further details. 

Going beyond pairwise feature interactions, many are interested in understanding how more complex, higher-order, and non-linear feature patterns contribute to a model's predictions. This growing area is often called representation learning and utilizes deep learning models like transformers to encode complex feature relationships in an often lower-dimensional representation space \citep{bengio2013representation}.  
While many of these feature representations are not directly interpretable, there is an active area of research to learn interpretable feature representations, especially in computer vision \citep{bengio2013representation}.

\subsubsection{Influential Points}

We have discussed interpretations of features in supervised models, but one can also interpret the observations through influential points, defined as observations whose removal significantly changes a model's prediction. There are a few model-specific approaches that provide intrinsic interpretations of influential points, like support vector machines, but most use model-agnostic strategies to identify these points. Coming from classical statistics, one can use strategies to detect outliers as well as measure the effect of removing each single training point \citep{hodge2004survey}. But more recently in machine learning, may have proposed using the influence function to approximate parameter changes for individual points based on the change in the gradient; these approaches have found widespread application in deep learning models \citep{koh2017understanding}.  

\section{Validating IML Discoveries}\label{sec:validation}
Interpretable Machine Learning techniques are being deployed across science and beyond to generate new knowledge or make data-driven-discoveries.  Yet one may ask, is my discovery true?  Or, have I discovered an artifact?  How can I tell the difference?  In other words, how can we validate discoveries made via Interpretable Machine Learning?  While most research in the Interpretable Machine Learning community has focused on developing new interpretability techniques, there has been relatively little work on the critically important problem of validation.  We contend that {\bf validation is one of the grand challenges in interpretable machine learning}, and this is especially crucial for making replicable, reliable, and trustworthy data-driven discoveries.  In this section, we motivate the necessity of validation for IML, discuss why this is so challenging, and then discuss several practical approaches that can be deployed with most IML techniques to help validate discoveries; we conclude with recommendations for validating IML discoveries in practice.

\subsection{Motivation \& Challenges}

\subsubsection{Motivation: Replicability, Reliability \& Trust}

Interpretable machine learning techniques are designed to always produce the desired interpretation, regardless of whether that interpretation or discovery truly reflects the underlying structure of the data. For example, $K$-means clustering always returns $K$ clusters whether there are groups in the data or not; feature selection always returns a subset of features whether the underlying true model is sparse or not.  Then, how can we tell if the machine learning interpretation generated a true discovery or is just an artifact in the data?  Further, there are a plethora of IML techniques and often each technique produces a different interpretation. Then, which interpretation is correct and represents a true discovery?  These are perhaps unknowable, epistemological questions.  Science addresses this by continually replicating and validating discoveries in follow-up studies until findings converge upon an accepted truth. Indeed, reproducibility and replicability are cornerstones of science \citep{national2019reproducibility,stodden2020theme}.

In machine learning, reproducibility means being able to obtain the exact same results after the same computational steps are performed on the same data, which is purely a computational concept \citep{willis2020trust,fineberg2020highlights} and is a prerequisite for validation.  Replicability means being able to obtain very similar results when two independent studies are performed to answer the same scientific question; in machine learning, this could entail performing the same or similar analysis on a new data set \citep{meng2020reproducibility,fineberg2020highlights}.
Replicability by itself, however, can not be the ultimate goal of scientific discoveries, as replicable results can still be wrong if the same mistakes are made in follow-up studies. 
Thus, going one step beyond this, many have advocated for reliability in machine learning, saying that predictions and findings should be robust to reasonable sensitivity tests like small changes in the data or the model, out-of-sample prediction tests, and consistency with domain knowledge \citep{meng2020reproducibility}. Validation for machine learning directly seeks to assess the replicability and reliability of results. For predictive tasks, there are well-developed and routinely employed validation strategies like data-splitting and cross-validation. For machine learning interpretations, however, there are very few widely accepted validation strategies and most employ IML techniques without any validation whatsoever in practice. For some uses of IML, 
this practice might not be terrible, but for the task of generating new discoveries, lack of validation is extremely damaging and could lead to erroneous, irreplicable and unreliable findings.  Indeed, there has been much commentary over the past several decades about a reproducibility and replicability crisis in science \citep{baker2016reproducibility}. Recently, several have suggested that failures to validate machine learning findings could be contributing to this crisis \citep{beam2020challenges,mcdermott2021reproducibility,gibney2022ai}. Validation is a crucial component of IML for generating data-driven discoveries.  

Beyond just the goal of utilizing best practices in science, replicability and reliability are critical to promote trust in machine learning results.  Many have lamented a lack of trust in machine learning systems and recommended to promote trust and societal acceptance of machine learning results by generating understandable interpretations \citep{jacovi2021formalizing,toreini2020relationship}. But, can we trust these interpretations? If interpretations are not replicable and reliable, then trust in these interpretations and discoveries breaks down. Recently, \cite{broderick2023toward} discussed these issues, among others, that cause trust to break down in probabilistic machine learning.  Further, they and several others have proposed various ways to enhance trust in machine learning results, emphasizing the need for validation strategies \citep{rasheed2022explainable,toreini2020relationship,broderick2023toward}.

\subsubsection{Challenges}\label{sec:validation_challenges}

To better understand why validating machine learning interpretations and their data-driven discoveries is so challenging, let us first discuss why a discovery might fail to validate. First, the machine learning model could be a poor fit to the data and hence any resulting interpretations would poorly reflect the signal in the data. Next, even if the model fits the data well, the interpretation approach could be a poor fit for the model, resulting in problematic interpretations; this can especially be the case with some post-hoc interpretability methods that fit a second model to generate the interpretation (e.g. LIME) \citep{molnar2020interpretable}. In addition, there could be a mismatch between the employed interpretation technique and the desired discovery task. For example, one important application of feature selection is to discover important biomarkers from high dimensional genomics data which is known to be highly correlated. Many feature selection techniques such as the Lasso are known to only select one feature out of a correlated set \citep{zou2005regularization} and hence would fail to identify some important but correlated biomarkers. Next, IML techniques are typically designed to find the desired interpretation in the data, regardless of whether that discovery truly exists in data. For example, $K$-means clustering will always discover $K$ clusters.  Many machine learning techniques are so powerful that they can always detect the rarest signals in large and complex data sets. For predictive tasks, we call this overfitting.  We also argue that {\it machine learning interpretations can be overfit to the training data}, and are hence challenging to validate.  

However, despite the fact that machine learning interpretations might fail to validate for a number of reasons, its validation is a significant challenge that has received surprisingly little attention in the literature \citep{rasheed2022explainable}. For prediction tasks, on the other hand, we have well-established techniques for validation: we ensure the predictive model generalizes to new, similar data. For instance, we can randomly split the available data into a training set which is used for building the predictive model, and a test set which is used for assessing the predictive accuracy of the model. 
Similar to that of predictive models, we say that {\em a machine learning interpretation validates if the resulting data-driven discovery generalizes well to new, similar data.}  Given this, one may ask: Can we simply employ a training and test set to validate interpretations?  We discuss this possibility subsequently, but in short, this prospect becomes much more complicated for interpretations.  First, many interpretable machine learning techniques are designed to make discoveries from the current (training) data but cannot directly apply the discovery to new data (e.g. clustering, manifold learning, feature importance ranking), and hence it is unclear how to assess how well it generalizes. Second, unlike the well-established prediction error metrics, there is no consensus on metrics for quantifying the accuracy of interpretations. Different machine learning interpretation approaches can yield very different discoveries and it is hard to come up with a fair metric that determines which one is the best on a test set. Most have suggested to assess machine learning interpretations via human evaluation of laypersons or domain experts \citep{carvalho2019machine,molnar2020interpretable,doshi2017towards}, but this does not lend itself to an easy-to-quantify metric analog of prediction accuracy.

\subsection{Practical Approaches for Validating Interpretations}\label{sec:validate_split}

In this section, we review two practical validation strategies that can be employed for almost any machine learning discovery.  While there may be additional validation approaches for specific interpretable machine learning models, we highlight these as they are fairly general and can be applied for both supervised and unsupervised discoveries.

\subsubsection{Data-Splitting}
As we previously discussed, randomly splitting the available data into a training and test set is the established mechanism for validating machine learning predictions.  Similar strategies can also help validate machine learning interpretations, but this also presents several challenges and limitations.  The key idea is to use IML on the training data to generate an interpretation as well as construct a prediction model based on this learned interpretation; then one can evaluate the model's predictive performance on the test data. For example, with projective dimension reduction techniques like PCA, one can learn the projection from the training set and then evaluate how the test data differs from its projection onto these components. Similarly in clustering, one could discover clusters on the training set, develop a classification model on the training set to discriminate these clusters, then apply this to the test set to predict cluster labels; one could then compare these predicted labels to those generated in an unsupervised manner by clustering the test set \citep{lange2004stability,handl2005computational}. 
For supervised discoveries like feature selection, one could discover important features on the training set as well as build a predictive model that only uses these features for the associated supervised learning task; then one can evaluate the prediction error of this model on the test set. The same idea can also be applied to feature interactions, feature patterns and other supervised discoveries.

Even though data-splitting provides a direct approach to validating IML discoveries, there remain many open questions and challenges.  Although we presented several examples of how this strategy can be used, it is unclear how to define an appropriate prediction task for some other machine learning interpretations, such as discovered associations and relationships between features, or anomalies and prototypes.  Next, this approach generates predictions on the test set, but what should these predictions be compared to?  For selecting important features, often the prediction error of a sparse model is not comparable to one with all the features.  Hence, how do we know whether the prediction error achieved by the smaller model is good enough to consider the feature subset validated?  Similar issues arise with many other IML approaches.  Finally, and arguably the most problematic aspect of data-splitting, the resulting interpretation is found using only part of the data and hence the interpretation might change with another randomly sampled training data.  This can be very troubling for replicability and in science.  Related, some would argue that since discovery is such a challenging task, one needs to use all available data and data-splitting reduces the amount of data available for the discovery stage.  

\subsubsection{Stability} 

\begin{figure}
    \centering
    \includegraphics[width=\textwidth]{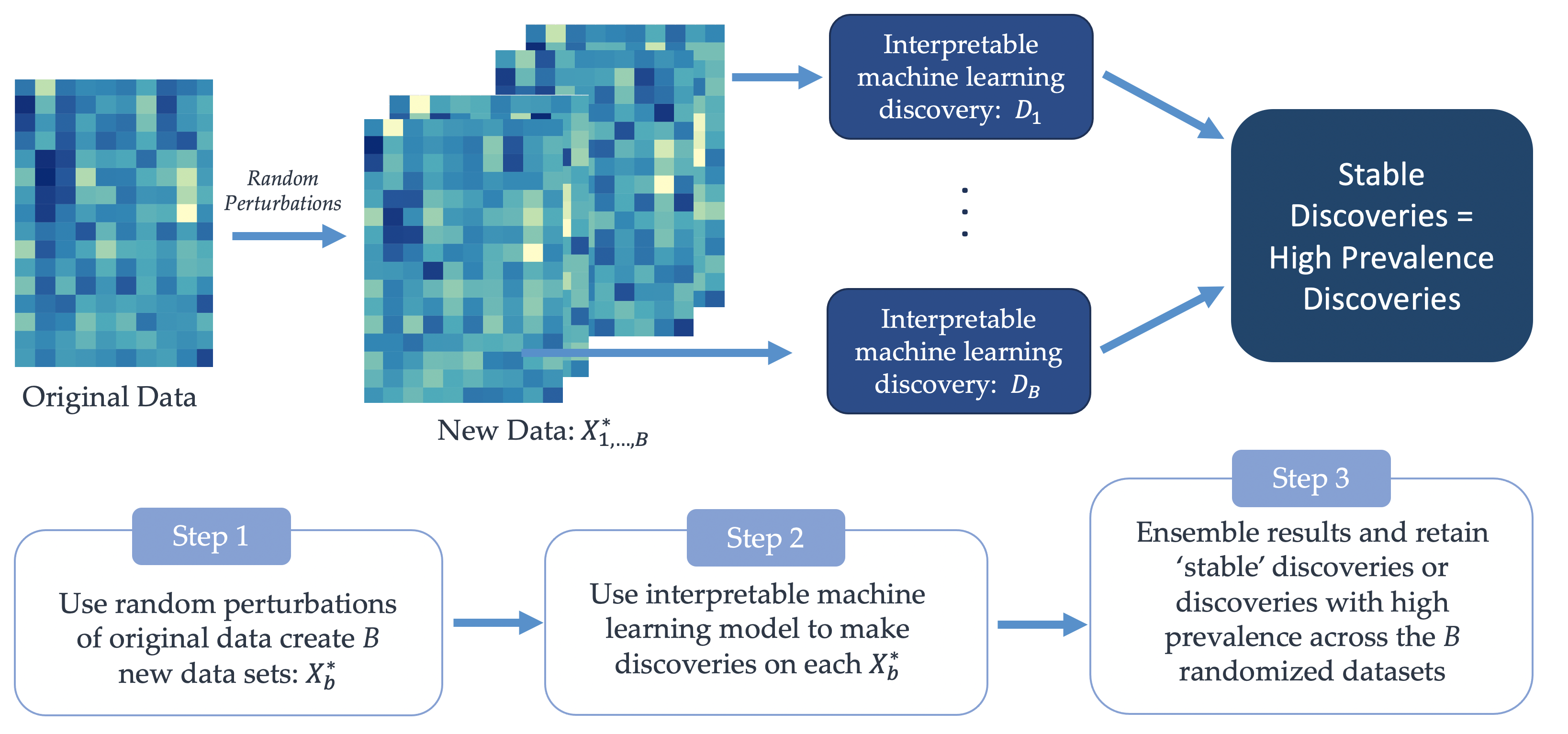}
    \caption{Illustration of stability principle for determining reliable data-driven discoveries.}
    \label{fig:stability}
\end{figure}

Another popular strategy for directly assessing the reliability of machine learning interpretations is the stability principle, which seeks to identify interpretations that are stable subject to random data perturbations. This idea was first introduced in \cite{meinshausen2010stability} in the context of feature selection with the Lasso. They randomly subsampled the data, fit a Lasso model to each subsample to select features, and the features that are selected most frequently over all subsamples are viewed as stable discoveries; importantly, they show that under certain assumptions this procedure controls the expected number of falsely selected features \citep{meinshausen2010stability}. There have also been many variants of this method studied in the statistical machine learning literature for feature selection \citep{shah2013variable}, feature interactions \citep{basu2018iterative}, graphical models \citep{liu2010stability}, PCA \citep{taeb2020false}, and clustering where it is commonly called consensus clustering \cite{monti2003consensus}.  Even though it has not been widely applied in other areas of machine learning, the idea of stability analysis is rather general and could be applied to any IML procedure.  We summarize the approach in Figure~\ref{fig:stability}.  First, the data is repeatedly randomly perturbed through subsampling, bootstrapping, randomly adding noise, or random data thinning \citep{neufeld2023data}. Then, IML procedures are used to make a discovery on each new random data set and discoveries with high frequency are declared as stable discoveries.  The core idea of stability is that discoveries that are not consistent under random data perturbations are more likely to be due to artifacts in the data or sampling noise, and hence are not replicable and reliable.  Indeed, stability analysis has received widespread attention, and many have advocated using this to validate discoveries and promote reproducibility in (data) science \citep{yu2020veridical}. It has also been widely used for solving many scientific problems such as discovering biomarkers in genomics \citep{he2010stable}. 

Despite the appeal of stability analysis to directly assesses reliability and overfitting, several challenges remain.  First, stability analysis can be computationally burdensome as it requires refitting the IML model many times; this is especially problematic for huge data or with complex models like deep learning.  Next, it is not always clear what type of random perturbation is appropriate and what quantitative criterion should by employed to determine stable discoveries for a given IML model and discovery task.  Consider consensus clustering where co-cluster membership is recorded for each subsample as there is not an easy way to record and ensemble the cluster membership.  Further, stability analysis could exacerbate mismatches between the interpretation techniques and the discovery task. Stability with the Lasso for feature selection, for example, is known to perform very poorly with correlated features. This is due to the fact that the Lasso may only select one among highly correlated features for each subsample, and hence none of these features would be deemed as stable, even if they are all important. Additionally, it is unclear whether the final stable discoveries are consistent with each other since they might not correspond to a single interpretation or set of interpretations from applying an IML model (e.g. stable features may not correspond to any Lasso solution at a single regularization parameter); many might consider this a disadvantage in scientific domains. Finally, and perhaps most importantly, stability analysis only assesses one form of reliability: the robustness of the discovery to small changes in the data, but not the robustness to changes in the modeling choice.  Furthermore and as we discussed earlier, reliability also means consistency with prior knowledge and out-of-sample predictive power, which are not reflected in stability analysis \citep{yu2020veridical}.  To see how this could be problematic, consider a scenario where a linear model is a poor fit for the data; then stability selection with the Lasso might result in stable features, but because of high model bias, these features will not be predictive or reflective of true important features.  Because of this, two different IML methods might lead to completely different discoveries that are each stable, which calls for further validation strategies.  To summarize, stability is necessary for indicating the reliability of a discovery, but it is not sufficient.

\subsubsection{Example: Validating Clusters}

\begin{figure}
    \centering
    \includegraphics[width=\textwidth]{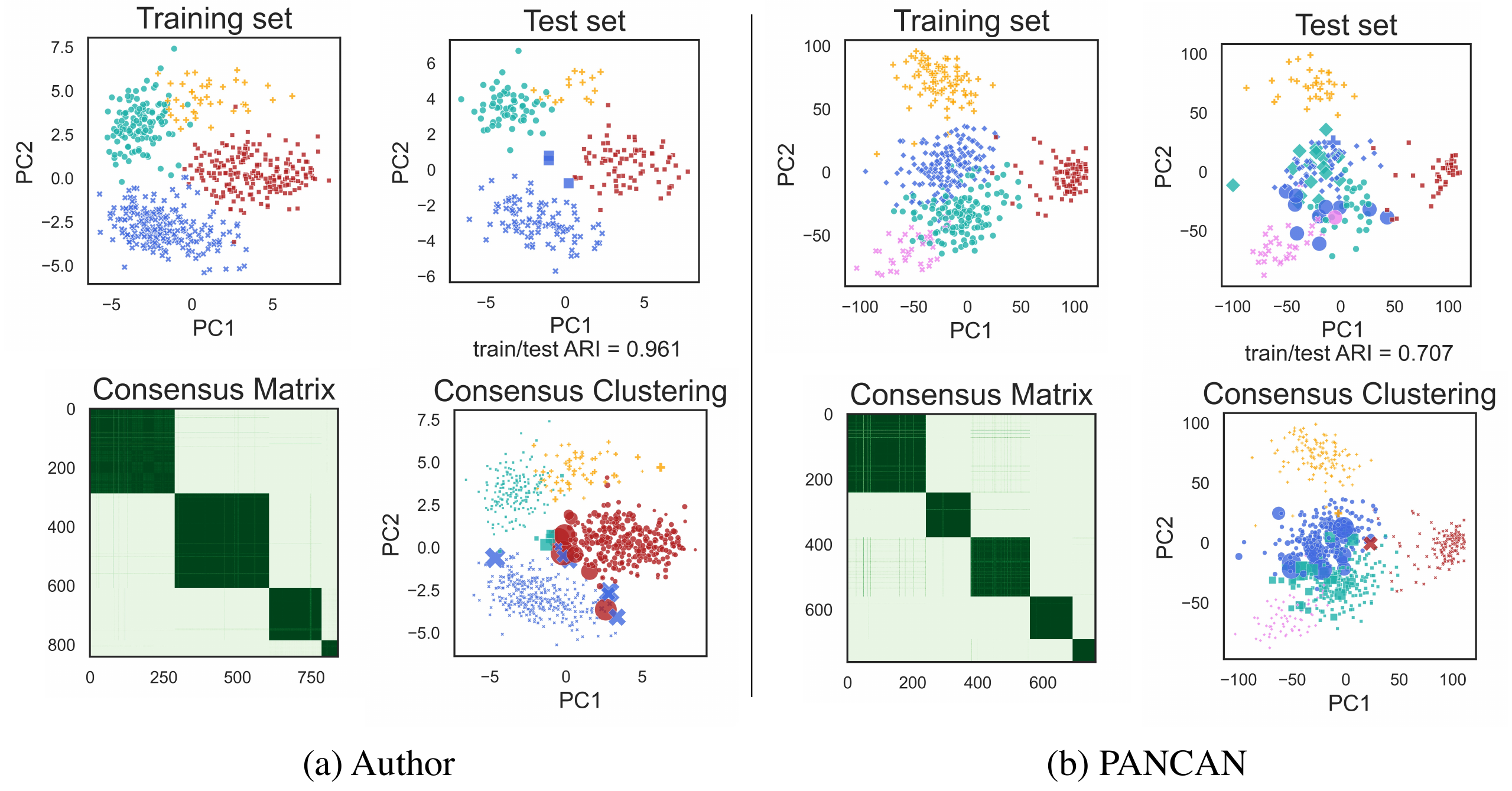}
    \caption{Example of how to validate cluster discoveries using data-splitting (top row) and the stability principle (bottom row) on two data sets. Results from the Author data set (a) show four well-validated clusters whereas results from the TCGA PANCAN data set (b) show that the blue and teal clusters are not well-separated and do not validate as well.}
    \label{fig:cluster}
\end{figure}

To illustrate data-splitting and stability analysis for validating IML, we turn to two real clustering examples: the Author data set with $n=841$ observations and $p=69$ features measuring the stop word count of book chapters from four English-language authors \citep{blake1998uci}, and the TCGA PANCAN data set with $n=761$ subjects and $p=13,244$ genes measuring the bulk RNA-seq gene expression for patients with five different types of tumors \citep{weinstein2013cancer}. We apply $K$-means clustering with $K=4$ and $K=5$ respectively, and seek to validate our discovered clusters using data-splitting and stability.  For data-splitting, we follow the predictive cluster validation approach outlined in \citep{lange2004stability,handl2005computational} by randomly taking 70\% of observations as a training set where we discover clusters as well as build a random forest classifier to predict these cluster labels.  We then independently cluster the remaining 30\% of observations in the test set as well as apply the random forest classifier to predict the labels; we measure the overlap between the predicted labels and the test set cluster labels using the adjusted rand index (ARI), a metric between zero and one with higher values indicating better cluster membership overlap. 
Results are shown in the top panel of Figure~\ref{fig:cluster} where we visualize the training and test set clusters in principal component (PC) scatterplots and highlight the test set observations where there is a mismatch between the predicted and cluster labels as larger points.  In the Author data, the training set predictions and test set cluster labels have a high degree of overlap, indicating strong validation of these four clusters.  Clusters in the PANCAN data set do not validate as well as shown by the lower ARI and the confusion of the cluster labels for the blue and teal clusters. 

Additionally, we apply the stability principle, which in clustering is often called consensus clustering \citep{monti2003consensus}, to validate these same cluster findings.  Specifically, we employ repeated data-splitting by repeatedly subsampling a training set on which we discover clusters and record the cluster co-membership; we then average these cluster co-memberships across all data splits to yield the $n \times n$ consensus matrix taking values between zero and one, with one indicating that the two observations were always assigned to the same cluster.  Heatmaps of the consensus matrix are shown in the bottom left panels of Figure~\ref{fig:cluster} with darker green indicating values at or near one.  We see that in both the Author and PANCAN data set, consensus clustering validates that there are $K=4$ and $K=5$ clusters respectively, exhibiting a clear block diagonal pattern.  Consensus clustering additionally allows us to inspect the uncertainty of cluster assignments for individual observations.  In the PC scatterplots on the bottom right of Figure~\ref{fig:cluster}, we show observations with point sizes inversely proportional to their cluster assignment uncertainty.  From this, we see that clusters in the Author data set are fairly well separated, but again, the teal and the blue clusters in the PANCAN data set exhibit a high degree of confusion.  Overall, both data-splitting and the stability principle can be used to validate cluster discoveries, and both of these methods reveal similar findings in the two examples we present.  

\subsubsection{Practical Recommendations}

We have discussed two general, practical validation strategies, but each of these has its own strength and limitations. Data-splitting can be a useful strategy for checking whether the discovered interpretation fits the data well, while stability analysis is most effective for evaluating whether the discovery is induced by random noise. Here, we would like to recommend one potential practical solution that leverages strengths of both data-splitting and stability analysis to evaluate the appropriateness of the chosen IML model (bias) and whether the ML technique overfits the noise (variance) in the IML discovery respectively. Specifically, one could employ repeated random data-splitting. On the training sets, one could apply the IML technique and record the resulting discovery as well as build a predictive model to predict that discovery.  On the test sets, one could evaluate the predictions and record the prediction error. Then, one could average both the discovery events and prediction errors across all randomized data sets. This is essentially repeated data splitting plus stability analysis, but importantly, allows one to assess both bias and variance associated with the IML procedure and discovery simultaneously. While we have not seen this technique explicitly outlined or employed, we have found this to be fruitful in our own work and thus recommend it as a possible validation strategy.  
Overall, validation is critically important for IML and especially for generating data-driven discoveries.  But the practical validation strategies we discussed are often constrained for certain types of discoveries; they often lack rigorous theoretical guarantees or quantitative guidelines on how to employ the approaches and which evaluation criterion to apply.  Hence, there are many opportunities for further research to develop and apply these practical validation strategies, study their theoretical properties, and determine the best validation approaches for many IML models and discovery types.

\section{Statistical Theory \& Inference for IML Discoveries}\label{sec:theory_inference}

Validation of IML discoveries is critical in practice to promote replicability, reliability and trust in data-driven discoveries.  But, theoretical guarantees and valid statistical inference offer a different perspective related to validation and are equally necessary to help build trust and promote replicability of IML discoveries \citep{rasheed2022explainable,broderick2023toward}.  In this section, we review statistical theoretical foundations for IML discoveries that address two pressing questions: (1) Under what data-generating models and under what conditions does an IML technique recover the true discovery with high probability? and (2) What is the uncertainty in a discovery or what discoveries can be trusted with a sufficient level of confidence?  We discuss these in Sections 5.1 and 5.2 respectively.

\subsection{Statistical Theory for Interpretable Machine Learning}\label{sec:theory}

Recently, \cite{broderick2023toward} argued that theoretical guarantees were important to help build trust in machine learning.  For IML discoveries, the goal is to theoretically characterize the type of data-generating models and the conditions under which IML techniques will make the desired discovery with high probability tending to one.  These types of theoretical guarantees largely fall under the areas of statistical consistency and selection consistency; the latter has received a huge amount of attention in the statistical machine learning community over the past two decades \citep{wainwright2019high}. Developing such theoretical foundations can help guide practitioners to choose the appropriate technique for their application and desired discovery task, understand when certain IML techniques will perform well and when they will not, and perhaps inspire the development of new IML techniques with improved performance and theoretical guarantees.  

Statistical consistency and selection consistency are well-studied for certain types of statistical and machine learning models, but are perhaps not readily applicable to other classes of machine learning methods leaving a gap in our theoretical understanding of IML. For example, classical statistical theory addresses the conditions under which parametric models like linear and generalized linear models consistently estimate their coefficients, a measure of intrinsic feature importance, in asymptotic and low-dimensional settings.  More recently, there has been a surge of interest in studying regularized versions of these and other semi-parametric statistical machine learning models in finite-sample and high-dimensional settings \citep{buhlmann2011statistics,wainwright2019high}.  Perhaps the most widely studied has been the Lasso, or $\ell_{1}$-regularized regression, for the IML task of feature selection \citep{tibshirani1996regression}.  For example, it is well established that the Lasso achieves selection consistency, or correct selection of true features with high-probability, under sparse linear regression models, when there is sufficient sample size relative to the log number of features, when there is sufficient signal in the true features, and under conditions like the irrepresentable, restricted eignenvalue, or incoherence conditions that limit the amount of correlation between features in the model \cite{zhao2006model}.  Consistency and selection consistency has also been established for many extensions of the Lasso, the Lasso in classification, semi-parametric models and other settings, and for other sparse regularizers \citep{buhlmann2011statistics}.  
Beyond feature selection and feature importance, several other model-specific and intrinsically interpretable unsupervised IML techniques have been studied theoretically under high-dimensional regimes. These include statistical consistency guarantees for clustering under a Gaussian mixture model \citep{loffler2021optimality}, network clustering under the stochastic block model \citep{abbe2017community}, low-rank estimation via PCA under spiked covariance models \citep{johnstone2009consistency}, and graph selection or structural graph learning for both Markov Networks (undirected graphs) and Directed Acyclic Graphs for causal discovery \citep{drton2017structure}; we refer the reader to \cite{wainwright2019high} for more details on many of these recent advances in high-dimensional statistical theory.

These advances in statistical theory provide assurance and insights for certain IML discoveries and certain techniques, but there remain are many limitations of this type of theory as well as open questions.  First, this statistical theory assumes a true population model that generates the data.  In practice, the true data generating process is unknown and uncheckable; it is often unclear how these IML techniques perform with mis-specified models.  Second, this type of theory is only applicable to model-specific and intrinsically interpretable IML techniques, which are often limited to linear or additive parametric or semi-parametric models.  Thus, this theory does not help us understand the performance of more flexible, non-linear modeling strategies like tree-based ensembles and deep learning.  Next, even when this type of theory is applicable to a particular model, techniques, and discovery task, the assumptions required to make the correct discovery with high probability are often hard to interpret and impossible to check in practice.  For example, it is impossible to check the irrepresentable condition \citep{zhao2006model} necessary for selection consistency of the Lasso for a particular data set without knowing the true features. Thus while this theory helps us understand the properties of certain IML techniques, this is unhelpful for trying to assess the validity of a particular discovery made by an IML technique on a particular data set.  Finally, statistical theory is currently very limited for interpretations of tree-based ensembles like random forests and boosting, neural networks and deep learning, and model-agnostic interpretations like Shapley values for feature importance.  Such areas provide many open research opportunities that would help us better understand these popular IML approaches and further promote trust in their discoveries \citep{broderick2023toward}.

\subsection{Statistical Inference for Interpretable Machine Learning}\label{sec:inference}

While statistical theory highlights the assumptions required to make an accurate discovery with high probability, another approach to validate discoveries is through statistical inference which quantifies the uncertainty associated with the discovery. Uncertainty quantification, typically through confidence intervals and hypothesis testing, is crucial in discerning whether a discovered pattern is due to random chance or is a genuine discovery.  This is especially important in high-stakes applications of IML where making decisions based on discoveries with a high degree of uncertainty could have devastating consequences; in science, this could lead to wasted resources and irreplicable results.  While uncertainty quantification for IML is a critically important task, this presents many challenges.  Note that the statistical theory discussed previously, in Section~\ref{sec:theory}, also quantifies errors for a discovery, but these cannot readily be used for uncertainty quantification as they depend on unknown parameters.  Similarly, practical validation approaches like data-splitting and stability, discussed in Section~\ref{sec:validation}, give a sense of the uncertainty in a discovery but cannot always be translated into rigorous statistical uncertainty quantification.  Nonetheless, uncertainty quantification has been studied for many of the same statistical machine learning models and IML tasks for which statistical theory has been developed; and more recently, uncertainty quantification has been considered in model-agnostic settings for specific IML tasks like feature importance and feature selection.  We briefly review these approaches.

Most statistical inference procedures are designed for model specific, global, and intrinsically interpretable statistical models that cover only a narrow range of IML techniques.  Classical inference approaches, which are typically asymptotic in nature, can be used for linear, generalized linear, or additive parametric (sometimes semi-parametric) statistical models to quantify the uncertainty in parameters.  Non-parametric or semi-parametric methods like bootstrap uncertainty quantification can also be used for many of the same methods.  More recently, several have developed inferential procedures for regularization techniques like the Lasso in high-dimensional regimes.  Such approaches include debiasing techniques \citep{van2014asymptotically}, which calculate the high-dimensional asymptotic distribution of the Lasso, and selective inference \citep{taylor2015statistical}, which computes confidence intervals and tests conditional on the Lasso solution.   Several others have recently employed similar strategies to quantify the uncertainty for unsupervised statistical learning tasks like clustering \citep{gao2022selective}, graphical models \citep{liu2013gaussian}, and principal components analysis \citep{koltchinskii2016asymptotics}.  This recent research on statistical inference for popular statistical machine learning models in high-dimensional regimes represents important advances in the field, but the approaches also have several limitations.  All of these approaches are model-specific and limited to parametric (or perhaps semi-parametric) statistical models, hence precluding application to popular non-linear machine learning models like kernels, tSNE, tree-based ensembles, and deep learning.  Further, these approaches assume the data arises from a specific generating model, which is not checkable in practice.  They are less effective at quantifying the uncertainty in IML discoveries when the model is mis-specified.

Given the limitations of model-specific approaches and following from recent developments in distribution-free predictive inference, many have advocated for model agnostic inference which can quantify the uncertainty associated with any IML model.  Thus far, such approaches have only been developed for feature importance and feature selection.   Among the first such approaches were based on the model-X knock-off framework which generates knock-off features with no relation to the response but that still retain the dependencies structure amongst the features \citep{candes2018panning,barber2019knockoff}.  This approach has been used to select features with false discovery rate control \citep{barber2019knockoff}, conduct conditional independence testing \citep{berrett2020conditional}, and construct confidence intervals for feature importance \citep{zhang2020floodgate}, among others.  The fact that knock-off approaches can be employed for any IML model is a major advantage, but this comes at the expense of assuming that the distribution of the features is known or can be closely approximated, a significant limitation in many domains.  Others have recently developed model agnostic inference approaches for feature importance; some consider feature occlusion inference \citep{lei2018distribution,williamson2021general,gan2022inference} which examines the prediction loss when removing one feature, while others consider the feature permutation test \citep{berrett2020conditional,kim2021local} which randomly permutes the feature of interest.  While very general and widely applicable, these approaches either perform inference for a random quantity that depends on the training set or require limiting assumptions on the data distribution or the consistency of the model employed.  In fact, the fundamental difficulty of distribution-free and model-agnostic feature importance inference has been recently revealed by \cite{shah2020hardness}, who shows that any conditional independence test that is valid without further assumptions on the data distribution or the model has no statistical power.  Hence while model-agnostic inference and uncertainty quantification for IML is critically important for validating many popular IML models, further research is needed to understand and work around limiting distributional and modeling assumptions.  Finally, there are many research opportunities to develop model agnostic inference approaches for IML tasks beyond feature selection and importance.

While this review has not focused on Bayesian machine learning, it is important to mention these techniques in the context of uncertainty quantification.  Indeed, many argue that one of the more appealing aspects of Bayesian approaches is the built-in uncertainty quantification through computing the posterior distribution and credible intervals; such approaches have been developed for IML tasks like feature importance and selection, graphical models, factor models, clustering, and more \citep{cortes2017bayesian,vallejos2015basics}.  Despite appealing uncertainty quantification properties, there are several challenges when applying these techniques to generate and validate IML discoveries.  First, computing or sampling from the exact posterior distribution is typically intractable or computationally prohibitive in big data settings.  Thus, people typically employ approximation techniques like variational inference \citep{blei2017variational}, but there is limited theory on how well these approaches work and how they affect the uncertainty quantification of IML discoveries.  Further, the IML discovery, posterior distribution, and any uncertainty quantification depend strongly on the prior employed.  The posterior distribution does not reflect this sensitivity to the prior and hence can underestimate the true uncertainty in the IML discovery; further sensitivity tests and model checking are needed for validation \citep{gelman2013philosophy,kruschke2021bayesian}.  We refer the reader to \citep{van2021bayesian} for more details.

In summary, statistical inference for IML discoveries is critical for validation and a growing area of research.  There are a number of important recent results in this field, especially for model agnostic inference, but there are also many open questions and challenges that are ripe areas for further research.

\section{Discussion}\label{sec:disc}

In this paper, we provided an overview of IML techniques that can be used for data-driven-discovery and discussed associated challenges and opportunities like validation, statistical theory, and inference.  But importantly, there are many aspects that we did not cover in this review that warrant further coverage and discussion in other works.  This paper focused on fairly general machine learning tasks and techniques, but there are an abundance of techniques developed for specific areas and tasks like those in computer vision, natural language processing and large language models, and reinforcement learning \cite{glanois2021survey}, among several others.  Many of these IML techniques can be used for discoveries and also share similar validation challenges. Another important area that we only briefly covered but deserves its own careful consideration is causality, which includes interpretability via counterfactual explanations \cite{mothilal2020explaining}, causal inference from interventional studies, and causal discovery from observational data. The latter can be especially important in science for discovering causal mechanisms, but perhaps faces even more challenges when it comes to validation, theory, and uncertainty quantification. We also only briefly discussed Bayesian machine learning and its associated uncertainty quantification, but this growing area of research deserves further discussion in the context of IML for generating new discoveries.

In this paper, we reviewed and discussed the grand challenge of how to validate discoveries made using IML.  We specifically discussed three aspects of this grand challenge: (1) practical tools for validating interpretations, (2) theoretical foundations of major IML techniques, and (3) uncertainty quantification for machine learning interpretations.  We presented two major types of practical validation strategies, data-splitting and stability, but each of these has their own caveats and limitations.  Further research is needed to combine the strengths of both approaches, elucidate a theoretical basis for these approaches, or perhaps develop a connection with uncertainty quantification via inference.  Next, we have a strong theoretical understanding of only a limited number of IML techniques, mainly those that are intrinsic, global, and model-specific.  This does not include interpretations of popular machine learning methods like random forests and deep learning; further research is needed to not only explain the strong predictive performance of these approaches but also understand their interpretations and discoveries.  Finally, there has been growing interest in uncertainty quantification for prediction, but quantifying the uncertainty of machine learning interpretations is also another critical component of validation that deserves further attention and research.  In addition to challenges associated with validation, there are also several other important questions that require further consideration and research.  Some of these include how to match the appropriate IML technique to the desired discovery task, how to compare different interpretations from different IML techniques, and how to marry domain knowledge and expertise with IML to better develop, deploy, and evaluate IML discoveries.

In summary, IML techniques hold great promise for making breakthroughs in science and beyond by mining ever larger data sets to detect the rarest signals.  But at the same time, these IML discoveries should be interpreted with caution without careful validation or uncertainty quantification.  Solving this grand challenge is critical for promoting replicable and reliable (data) science as well as trustworthy machine learning; they also provide exciting research opportunities at the intersection of statistics and machine learning.

\bibliographystyle{authordate4}

\bibliography{interpretability.bib}

\end{document}